\documentclass[preprint,12pt]{elsarticle}

\usepackage{amssymb}
\usepackage{amsmath} 
\usepackage{graphicx}
\usepackage{soul}
\usepackage{float}
\usepackage{xcolor}

\usepackage{booktabs} 
\usepackage{threeparttable} 

\begin{document}

\begin{frontmatter}



\title{DG-STMTL: A Novel Graph Convolutional Network for Multi-Task Spatio-Temporal Traffic Forecasting}


\author[inst1]{Wanna Cui}

\affiliation[inst1]{organization={Department of Engineering, King's College London},
            addressline={Strand}, 
            city={London},
            postcode={WC2R 2LS}, 
            country={United Kingdom}}

\author[inst2]{Peizheng Wang}
\author[inst1]{Faliang Yin}

\affiliation[inst2]{organization={Department of Civil and Environmental Engineering, Imperial College London},
            addressline={South Kensington Campus, Exhibition Road}, 
            city={London},
            postcode={SW7 2AZ,}, 
            country={United Kingdom}}

\journal{arXiv}
\begin{abstract}

Spatio-temporal traffic prediction is crucial in intelligent transportation systems. The key challenge of accurate prediction is how to model the complex spatio-temporal dependencies and adapt to the inherent dynamics in data. Traditional Graph Convolutional Networks (GCNs) often struggle with static adjacency matrices that introduce domain bias or learnable matrices that may be overfitting to specific patterns. This challenge becomes more complex when considering Multi-Task Learning (MTL). While MTL has the potential to enhance prediction accuracy through task synergies, it can also face significant hurdles due to task interference. To overcome these challenges, this study introduces a novel MTL framework, Dynamic Group-wise Spatio-Temporal Multi-Task Learning (DG-STMTL). DG-STMTL proposes a hybrid adjacency matrix generation module that combines static matrices with dynamic ones through a task-specific gating mechanism. We also introduce a group-wise GCN module to enhance the modelling capability of spatio-temporal dependencies. We conduct extensive experiments on two real-world datasets to evaluate our method. Results show that our method outperforms other state-of-the-arts, indicating its effectiveness and robustness.

\end{abstract}


\end{frontmatter}
\section{Introduction}

Accurate spatio-temporal prediction is important in domains like intelligent transportation system \cite{wang2020traffic}, \cite{zhang2020spatio}, \cite{yu2020forecasting}, \cite{son2023forecasting}. Such predictions are key to improving resource efficiency and decision-making. For example, by accurately predicting traffic patterns, intelligent traffic management systems can proactively adjust signal timings and manage traffic flow \cite{kong2013utn}, \cite{nallaperuma2019online}, \cite{papageorgiou2007its}. This could help in mitigating congestion before it becomes severe, ensuring smoother traffic movement across urban areas \cite{jia2017data}. Additionally, these systems can inform travelers about the best times to travel or alternative routes, improving the overall travel experience by reducing delays.

Accurate traffic forecasting remains is challenging as it requires a deep understanding of the complex interdependences within the data \cite{baggag2019learning}, \cite{james2021citywide}, \cite{zeb2024generalized}. Take urban traffic flow as an example: the flow in a specific city area is influenced not just by the time but also by nearby occurrences like road closures, accidents, or significant events such as concerts or sports games. These events create a ripple effect, changing traffic patterns not only locally but across the entire road network. Traditional spatio-temporal prediction approaches often involve various deep neural network , including Convolutional Neural Networks (CNNs) and Recurrent Neural Networks (RNNs) \cite{zhang2019flow}, \cite{zhang2018predicting}, \cite{zonoozi2018periodic}. However, these methods struggle with grid-based data, which is inadequate for capturing the irregular topology of real-world road networks \cite{ali2022exploiting}, \cite{li2023dynamic}. 
In contrast, graph-based data representations accommodate these features more naturally and flexibly, improving the modeling of complex relationships and topology.

Recently, researchers have shifted their attention to Graph Convolutional Network (GCN) as its capability in learning complex representations directly on graph \cite{zhang2023autostl}, \cite{bhatti2023deep}, \cite{khoshraftar2024survey}. Numerous models based on GCN have been proposed to model the complex spatio-temporal relationships inherent in data. However, challenges still exist that prevent fully leveraging GCNs for forecasting complex spatio-temporal dynamics. A significant challenge lies in defining the adjacency matrix. This matrix plays a key role in the GCN's modeling capabilities, as it determines how information is aggregated from a node's neighbors \cite{bikram2024attentive}. Most existing research uses static adjacency matrices based on established knowledge, such as the road connectivity, which do not adequately capture evolving spatial and temporal dynamics. \cite{geng2019spatiotemporal}, \cite{xu2024coupled}.While this approach is straightforward and offers a measure of stability, it fails to fully capture the spatial and temporal dynamics \cite{liu2020dynamic}. Both \cite{bai2020adaptive} and \cite{shi2023dagcrn} have identified this limitation, demonstrating through their experiments that static adjacency matrices do not adequately reflect the time-varying dynamics of spatial dependencies. To this end, several studies have adopted dynamic adjacency matrices that are learnable throughout the model training \cite{skarding2021foundations}, \cite{ye2022learning}. However, this method also comes with its challenges, such as the risk of overfitting to specific patterns in the data. The challenge above becomes more complex within a spatio-temporal multi-task framework, where multiple tasks involving spatial and temporal data are solved simultaneously. This complexity, however, presents an opportunity, through the advantages of Multi-Task Learning (MTL), which offers advantages in tackling such predictive tasks. Traditional models, primarily designed for single-task forecasting \cite{zhang2021traffic}, \cite{zhou2021ast}, often overlook the potential benefits from the synergistic effects in MTL. By effectively leveraging shared information across related tasks, MTL can significantly improve the predictive accuracy of each individual task \cite{zhang2023autostl}. For instance, in urban ride-hailing services, the demand for pick-ups and drop-offs forms an interconnected system, especially in high-traffic areas like shopping malls. However, this approach may also lead to task interference, a key challenge in MTL, where the objectives of different tasks may conflict. Consider a model trained for both predicting traffic flow and speed. Optimizing for speed prediction might require the model to focus more on short-term fluctuations influenced by factors like traffic signals or temporary blockages. In contrast, flow prediction might benefit from understanding longer-term patterns influenced by daily commuting patterns. This highlights the need for a MTL approach capable of efficiently sharing information between tasks while minimizing their negative interference.

To solve the challenges above, this paper proposes a novel framework: Dynamic Group-wise Spatio-Temporal Multi-Task Learning (DG-STMTL). In contrast to previous approaches that rely on either static adjacency matrices or entirely learnable ones, our framework proposes a hybrid adjacency matrix generation strategy. This approach effectively balances the stability of static matrices with the adaptability of dynamic ones into an unified matrix, enabling more effective representation. Additionally, to keep the integrity of single-task learning while promoting positive synergistic interactions among tasks, we introduce two information sharing mechanisms. For modeling spatio-temporal dependencies, we design a novel GCN architecture. This design adopts a two-stage grouping strategy: initially, it groups data according to temporal proximity, focusing on short-term dynamics. Then, a feature-level grouping is applied, partitioning the data into clusters for specific convolution operations, effectively capturing long-term dependencies. This hierarchical process enables a deeper and more granular analysis of spatio-temporal relationships, enhancing the model's ability to predict complex patterns. The key contributions of our study are summarized as follows: 

\begin{itemize}
    \item We propose DG-STMTL, a novel framework for enhancing the spatio-temporal multi-task learning capability. Unlike traditional models, our framework effectively mitigates task interference while promoting the positive collaborative effects of MTL through efficient information sharing. This ensures improved accuracy and robustness in complex spatio-temporal prediction tasks.

    \item We introduce a novel adjacency matrix generation module that combines the stability of static adjacency matrices with the adaptability of dynamic matrices. This hybrid approach allows the model to better capture and respond to the evolving dynamics of real-world data, addressing the limitations of existing models that rely solely on static or dynamic matrices.

    \item We develop a GCN-based spatio-temporal learning module with a group-wise approach, effectively capturing complex dependencies within and across groups. 
    
    \item We conduct extensive experiments on two real-world datasets to validate the proposed framework. The results demonstrate that our DG-STML not only outperforms the performance of existing state-of-the-art models but also maintains robustness across a range of spatio-temporal data scenarios.
\end{itemize}

The structure of the rest of the paper is organized as follows: Section 2 reviews related work in spatio-temporal prediction. Section 3 introduces the preliminary concepts necessary for understanding the proposed approach. Section 4 describes the DG-STML framework in detail. Section 5 reports on the experimental results and provides a thorough analysis. Finally, section 6 concludes the paper.

\section{Related Work}

\subsection{Traditional Spatio-Temporal Prediction Methods}
Traditional prediction methods have mainly used time series analysis and machine learning models \cite{koutsaki2023spatiotemporal}. Time series models such as Autoregressive Integrated Moving Average (ARIMA) and its several variants have received particular attention due to its capability to handle non-stationary time series \cite{chen1995analysis}. While ARIMA models excel in managing time series data, they have limitations in handling spatial information and complex spatio-temporal patterns, which can limit their predictive accuracy in dynamic and heterogeneous environments. Machine learning methods, such as Support Vector Machines (SVM) \cite{noble2006support} and decision trees \cite{rokach2005decision}, can capture more complex relationships but often require extensive data and feature engineering. Additionally, these methods may not effectively integrate spatial and temporal data simultaneously, which can reduce their effectiveness in spatio-temporal prediction tasks\cite{koutsaki2023spatiotemporal}.

In recent years, there has been a growing trend in deep learning models for spatio-temporal prediction. This includes deep neural network architectures such as RNNs \cite{yang2019traffic}, which is highly effective in modeling the temporal dependencies. In contrast, CNNs \cite{o2015introduction} based models demonstrate superior spatial modeling capabilities by treating input data as an image. In such settings, each data point is treated as a pixel, and convolution operators are employed to effectively extract local spatial features. These features are then combined to form higher-order representations. Furthermore, some hybrid models were developed such as ConvLSTM \cite{shi2015convolutional}, which combines the spatial modeling capabilities of CNNs with the RNNs. A common limitation across these approaches is that they treat spatio-temporal data in grid-based format. However, real-world data can exhibit irregularities and complex interdependencies. Such structures are not always be captured by traditional convolutional or recurrent networks.

\subsection{Graph Based Spatio-Temporal Prediction Methods}

GCN emerges as a promising technique to address the above limitation by operating directly on graph structure, which are often natural representations for many spatio-temporal datasets \cite{su2022convolutions}. GCNs are specifically designed to handle non-Euclidean structures, enabling them to efficiently manage irregular spatial patterns \cite{zhang2019graph}.

Recent advancements in spatio-temporal prediction have integrated RNNs with GCNs. For instance, the Spatio-Temporal Graph Convolutional Network (ST-GCN) \cite{yu2017spatio} combines the spatial modeling strengths of GCNs with the temporal modeling capabilities of RNNs, effectively capturing both spatial and temporal dynamics. Similarly, the Diffusion Convolutional Recurrent Network (DCRNN) \cite{li2017diffusion} addresses traffic flow forecasting by modeling spatial and temporal dependencies through a diffusion process across urban road networks. The Graph Convolutional Recurrent Network (GCRN) \cite{seo2018structured} combines the spatial convolution features of ChebNet with the temporal dynamics of LSTM, enabling robust modeling of spatio-temporal data. Similarly, the Diffusion Convolutional Recurrent Network (DCRNN) \cite{li2017diffusion} addresses traffic flow forecasting by modeling spatial and temporal dependencies through a diffusion process across urban road networks. Despite these advancements, several research gaps remain. First, the iterative nature of RNN propagation introduces complexity, often resulting in information loss and gradient issues. Therefore, it is necessary to develop more efficient GCN-based model architectures to effectively learn complex spatio-temporal dependencies. Moreover, the adaptability of these models to dynamic and evolving spatio-temporal patterns is limited, necessitating the development of more robust and flexible frameworks.

In response to the limitations of previous methods, new models such as the Spatial-Temporal Synchronous Graph Convolutional Network (STSGCN) \cite{song2020spatial} and the Automated Dilated Spatio-Temporal Synchronous Graph (Auto-DSTSG) \cite{jin2022automated} have been introduced. These models leverage spatio-temporal synchronous graphs for modeling relationships in a unified framework and capturing both localized and global spatio-temporal dependencies.
Auto-DSTSG, in particular, opens new way for adjacency matrix formulation by integrating graph structure search techniques, enabling the automatic identification of the optimal adjacency matrix configuration. Despite these advancements, both STSGCN and Auto-DSTSG rely on predefined adjacency matrices based on prior domain knowledge. This approach can introduce domain bias and limit the models' ability to capture short-term dynamic relationships, potentially reducing their adaptability to changing spatio-temporal patterns. To address these issues, the Adaptive Fused Spatial-Temporal Graph Convolutional Network (AFSTGCN) \cite{xiao2022afstgcn} introduces a learnable approach to generating the spatio-temporal adjacency matrix. AFSTGCN uses randomly initialized parameters that are adapted during training to create an adaptive adjacency matrix. While this method provides greater flexibility and can potentially improve adaptability, it often struggles with overfitting to historical data patterns. This overfitting can make it challenging for the model to adjust to new or changing conditions, which may limit its effectiveness in real-world scenarios. Therefore, there is a need for more robust frameworks that can efficiently balance flexibility and robustness, ensuring accurate spatio-temporal predictions across varying conditions. Addressing these gaps will be crucial for advancing the field of spatio-temporal modeling.

\subsection{Spatio-Temporal Multi-Task Learning}

Recently, the research community has shown a growing interest in Spatio-Temporal Multi-Task Learning (STMTL), a variant of MTL that focuses on leveraging spatio-temporal correlations. STMTL aims to enhance single-task models by simultaneously addressing multiple tasks within spatio-temporal data. In current STMTL approaches, a common strategy is to use a shared layer, usually a fully-connected layer, to integrate various tasks. This shared layer is crucial for extracting valuable information across different tasks. After this integration, each task is then individually processed through its own task-specific learning modules \cite{zhang2023autostl}, \cite{feng2021multi}. Multi-View Multi-Task Spatio-Temporal Networks (MVMT-STN) employs GCN as a shared information layer for two tasks, enhancing individualized predictions while sharing features \cite{wang2021traffic}. A recent advance in STMTL is the Automated Spatio-Temporal Multi-task Learning (AutoSTL), which introduces a shared module to capture common knowledge and task correlations. This approach enhances multi-task spatio-temporal predictions by fostering collaborative learning across tasks \cite{zhang2023autostl}. Despite these advancements, the current STMTL frameworks often fail to efficiently share information between tasks, thus limiting the full potential of MTL. Hence, a primary aim of this paper is to develop a STMTL framework designed for efficient information exchange among tasks while simultaneously solving multiple tasks.

\section{Preliminaries}

\subsection{Problem Definition}

The problem of this study is the prediction of spatio-temporal multi-task traffic data such as traffic flow and speed. The spatio-temporal graph is denoted as \( G = (N, A) \), with \( N \) represents the set of nodes that correspond to distinct segments of the road network or urban region, and \( A \) denotes the adjacency matrix, which represents the relationship among the nodes. Given the spatio-temporal feature matrices \(X_1\),\(X_2\), \(\ldots\), \(X_k\) for task 1,2,..., \(k\) and the corresponding target matrix \( Y \), the goal is to learn a mapping function \( f(\cdot) \) for all tasks such that the overall loss function is minimized.

\begin{equation}
\min_{\mathcal{\theta}} \mathcal{L}(Y, f(X_1, X_2, \ldots, X_k; \mathcal{\theta}))
\quad 
\end{equation}

where \( \mathcal{\theta} \) denotes the parameters of the model, \( \mathcal{L} \) denotes the loss function specific to MTL, \( f(\cdot) \) represents the prediction model.

\subsection{Basic Adjacency Matrix Formulation}

This section defines several adjacency matrices, which are derived from human knowledge. These matrices are binary, where 1 and 0 denote the presence and absence of a specific type of relationship, respectively.

\paragraph{Physical connectivity matrix $A_{S}$} It represents the physical connections among nodes. The value 1 represents the presence a direct physical connection between nodes.

\paragraph{Temporal connectivity matrix $A_{T}$} It indicates the self-connection of nodes over time, represented by a diagonal filled with ones.

\paragraph{Spatio-temporal connectivity matrix $A_{ST}$} It represents connections among nodes based on similarities in their spatio-temporal patterns. To quantify these similarities, the Pearson correlation coefficient is employed with a threshold to control the sparsity

\section{Methodology}

\subsection{Overview}

In this section, we detail the proposed DG-STMTL. Figure 1 gives an overview of the overall architecture. It includes three main components: the Hybrid Adjacency Matrix Generation (HAMG) module, the Cross-Task Knowledge Exchange (CTKE) unit, and the Group-wise Spatio-Temporal Graph Convolutional (GSTGC) module. Specifically, the HAMG module is designed to generate hybrid adjacency matrix for multiple tasks, formed by a combination of a static adjacency matrix and a dynamic adjacency matrix. Within the HAMG module, We design the CTKE unit to dynamically generate the adjacency matrix from input spatio-temporal sequence. This component allows adaptive adjustments based on data dynamics and introduces a graph-based multi-task information sharing mechanism. The GSTGC module is designed to improve the modeling capability of complex spatio-temporal relationships. Integrating these modules, our framework significantly improves spatio-temporal modeling capability, flexibility, and efficiency in MTL. In the following sections, we will introduce the details of each component.

\begin{figure*}[ht]
    \centering
    \includegraphics[width=\textwidth]{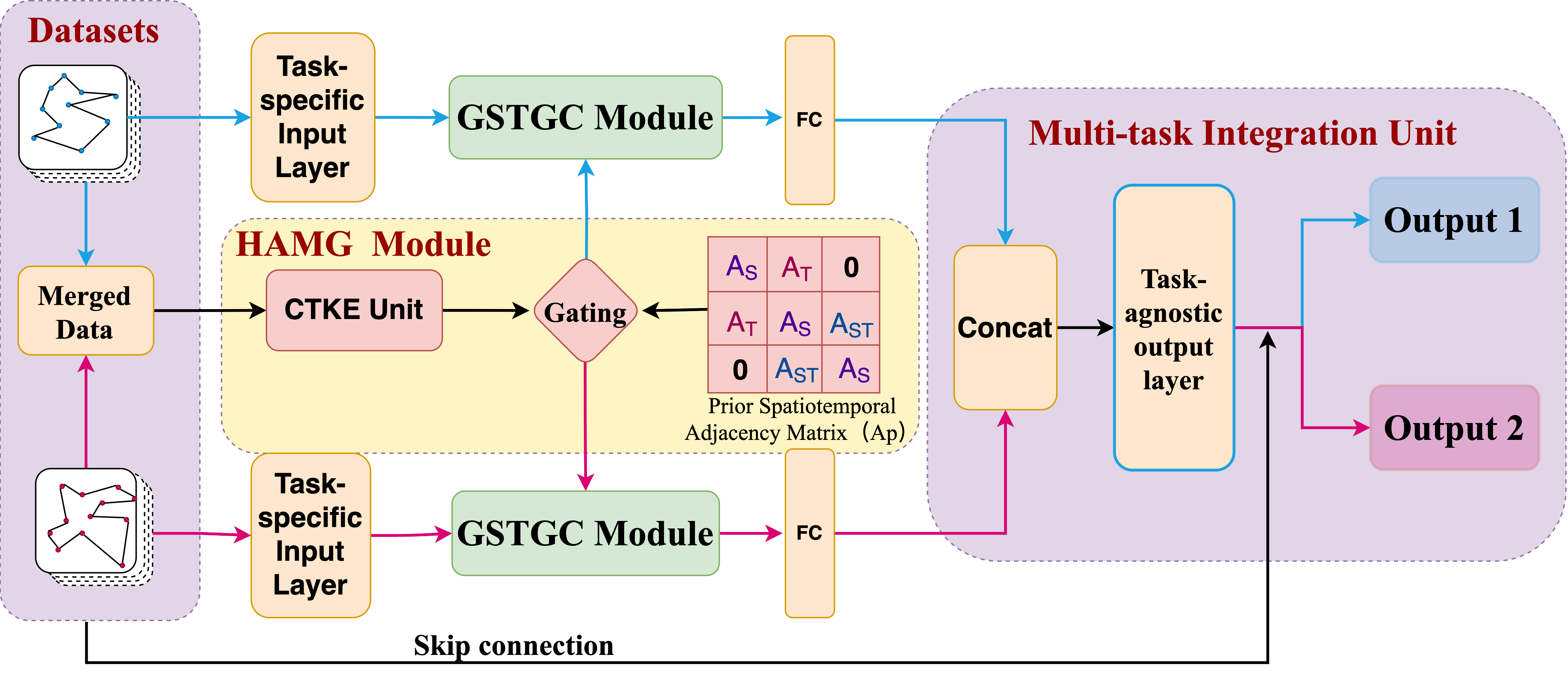}
    \caption{Overview of the Dynamic Graph-based Multi-Task Learning Framework Architecture}
    \label{fig:overview}
\end{figure*}

\subsection{Cross-Task Knowledge Exchange Unit}

Unlike the traditional MTL methods that use fully connected layers for information sharing, we design the CTKE unit as a novel graph-based mechanism for efficient information sharing. The CTKE aims to enhance learning capabilities across multiple tasks by constructing a shared adjacency matrix, dynamically generated from input data. This approach not only facilitates a more effective exchange of knowledge among tasks but also enhances adaptability to dynamic data. We first concatenate data from multiple tasks to form a unified multi-task feature matrix \(X\):

\begin{equation}
X = \text{Concatenate}([X_{1}, X_{2}, \ldots, X_{k}]) \in \mathbb{R}^{N \times T \times K \times C}
\end{equation}

where  \(N\) is the number of nodes, \(T\) is the length of the historical sequence,  \(K\) is the number of tasks, and \(C\) is the number of features associated with the task. To further enhance the representation power of the spatio-temporal data, we employ a broadcasting-based spatio-temporal multi-task embedding. Specifically, this mechanism incorporates two embedding matrices: temporal task embedding (TTE) \(E_{TK} \in \mathbb{R}^{1 \times T \times K \times 1}\) and spatial task embedding (STE) \(E_{SK} \in \mathbb{R}^{N \times 1 \times K \times 1}\) for the temporal dimension and the spatial dimension, respectively. These temporal and spatial embedding matrices are learnable throughout training and use broadcast operation, which can be formulated as follows:

\begin{equation}
    X_{\text{embedded}} = X + E_{TK} + E_{SK} \in \mathbb{R}^{N \times T \times K \times C}
\end{equation}

where \(X_{\text{embedded}}\) denotes the embedded feature matrix. By integrating TTE and STE, we enrich the information within the original multi-task data matrix. In our framework, an essential step is generating a dynamic adjacency matrix, denoted as \(B\), based on the embedded multi-task feature matrix. This process is crucial for effectively sharing information across multiple tasks. The adjacency matrix in this paper has dimension \(mN \times mN\), which differs from the conventional \(N \times N\) structure. This particular design aligns with the structure of the static adjacency matrix employed in this paper, known as the spatio-temporal synchronous adjacency matrix \cite{song2020spatial}. This method will be introduced in the next section. Starting with the embedded data \(X_{\text{embedded}}\), we apply a sequence of operations to extract inter-feature relationships. Specifically, we apply a max pooling to aggregate the embedded matrix across the temporal dimension \(T\). The aggregated matrix is then reshaped and processed through a fully connected layer to transform the feature dimension and integrate multi-task information. These operations can be summarized by the following formulas:

\begin{equation}
X_{\text{agg}} = \max_{t=1}^{T} X_{\text{embedded}}(:, t, :, :) \in \mathbb{R}^{N \times K \times C}
\end{equation}

\begin{equation}
X_{\text{transformed}} = \sigma(Reshape(X_{\text{agg}})W + b) \in \mathbb{R}^{N \times D}
\end{equation}

where \(W \in \mathbb{R}^{CK \times D}\)  and \(b \in \mathbb{R}^{D}\) are learnable parameters, and \(\sigma\) denotes the ReLU activation function. It should be noted that the transformed dimension \(D\) should be divisible by \(m\), where \(m\) represents the number of time steps considered by the adjacency matrix. To align with the scaling factor \(m\) and prepare for adjacency matrix construction, the obtained \(X_{\text{transformed}}\) is reshaped to dimension \(mN \times D'\), where \(D'\) = \(D/m\). The reshaped matrix is then used to construct the correlation matrix by calculating the dot product of the reshaped matrix with its transpose. This operation yields a square matrix of dimensions \(mN \times mN\). Finally, a softmax normalization is applied to the correlation matrix:

\begin{equation}
X_{\text{reshaped}} = \text{Reshape}(X_{\text{transformed}}) \in \mathbb{R}^{mN \times D'}
\end{equation}

\begin{equation}
    C = X_{\text{reshaped}} X_{\text{reshaped}}^{\top} \in \mathbb{R}^{mN \times mN}
\end{equation}

\begin{equation}
B = \text{softmax}(C) \in \mathbb{R}^{mN \times mN}
\end{equation}

To further clarify the design rationale, we provide additional insights into the formulation of the correlation matrix. First, the dot product effectively captures feature-level correlations between nodes while maintaining computational efficiency. Compared to alternatives such as attention-based mechanisms, which offer greater flexibility but introduce additional complexity and computational overhead, our design achieves a more favorable trade-off between efficiency, stability, and performance, making it particularly suitable for large-scale spatio-temporal tasks. Second, the Softmax normalization ensures that the generated adjacency matrix remains well-calibrated, with the resulting edge weights constrained to a normalized range. This not only avoids numerical instability but also facilitates smooth gradient propagation, contributing to stable and reliable learning dynamics.

\subsection{Hybrid Adjacency Matrix Generation Module}

In this section, we introduce the HAMG module for generating adjacency matrix by integrating a prior spatio-temporal adjacency matrix with its dynamic counterpart obtained from the CTKE unit via a gating mechanism. The generated adjacency matrix is designated as the input for the GSTGC module. The combination of static and dynamic adjacency matrices is motivated by their complementary strengths. Static matrices provide a stable prior structure derived from domain knowledge, ensuring robustness, while dynamic matrices adapt to evolving patterns in the data, capturing short-term dependencies. However, relying solely on either component can lead to limitations such as rigidity (static) or overfitting (dynamic). By integrating both through a task-specific gating mechanism, the hybrid adjacency matrix achieves a balance between stability and flexibility, allowing the model to adapt to task-specific dynamics while retaining generalization capability.

\paragraph{Prior Spatio-Temporal Adjacency Matrix Generation}

This paper employs a special design of adjacency matrix which has the dimension  \(mN \times mN\). Specifically, the prior spatio-temporal adjacency matrix \(\mathcal{A}_P\) integrates three foundational elements as introduced in the preliminary section: $A_{S}$, $A_{T}$,$A_{ST}$. The objective of this integration process is to capture the complex spatio-temporal interactions among nodes over \(m\) distinct time steps. In this paper, we set \(m\) = 3. The formal expression of this integration is given by:

\begin{equation}
\mathcal{A}_P\ = F(A_S, A_T, A_{ST}) \in \mathbb{R}^{3N \times 3N}
\end{equation}

where \( F(\cdot)\)  denotes the integration function that combines the three matrices, thereby offering a high flexibility in modeling spatio-temporal dependencies. Figure 2 illustrates several potential ways for combining these matrices. The generation of the prior spatio-temporal adjacency matrix is a structured process that synthesizes spatial, temporal, and spatio-temporal correlations among nodes to capture comprehensive dependencies. Specifically, the spatial matrix $A_{S}$ captures physically spatial connections within the same time step. The temporal matrix $A_{T}$ captures the evolution of nodes between consecutive time steps. This facilitates an understanding of temporal patterns and changes in node behavior over time. The $A_{ST}$ represents the spatio-temporal relationships between nodes across two time steps, capturing spatio-temporal patterns during these intervals. 

\begin{figure}[h]
    \centering
    \includegraphics[width=1\linewidth]{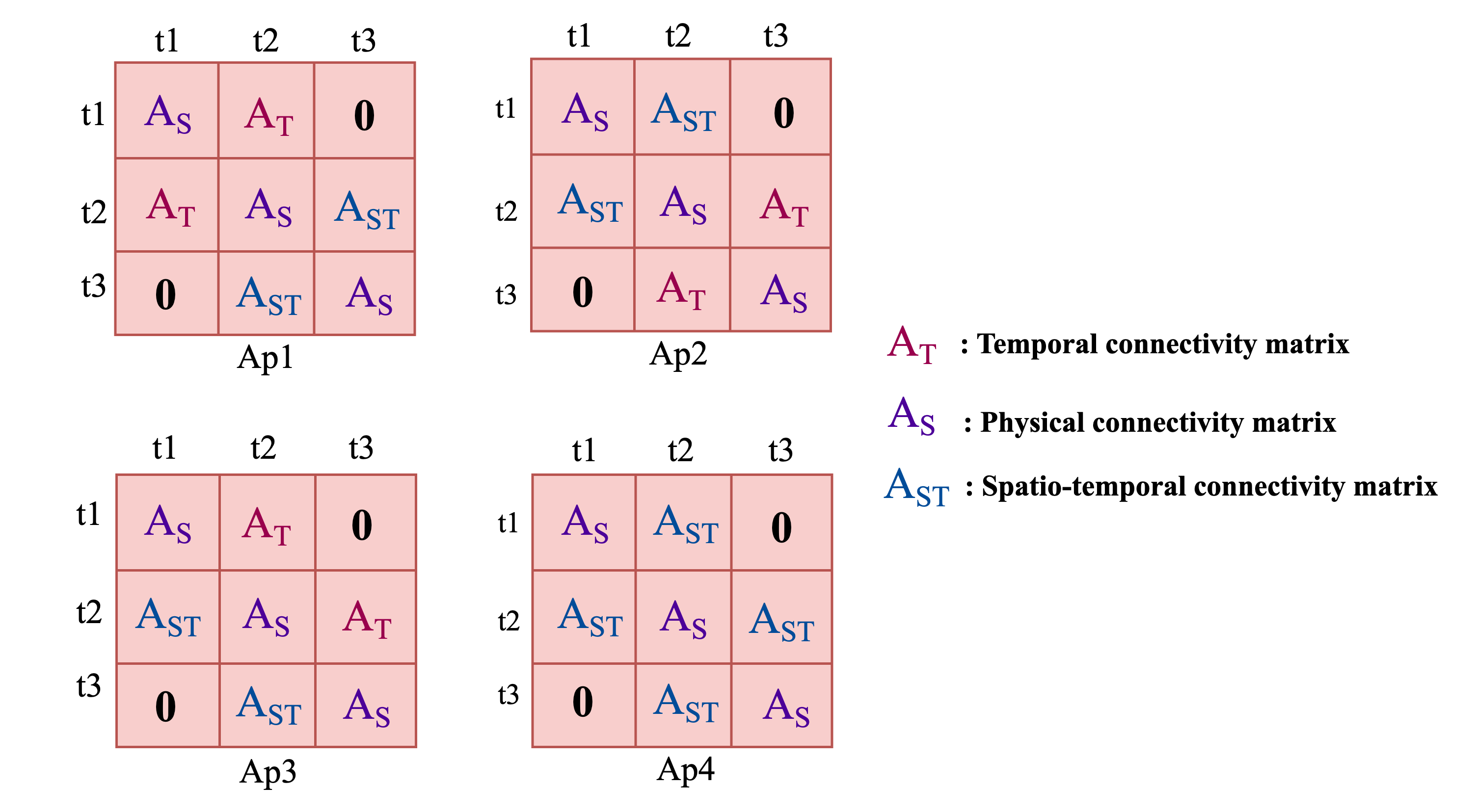}
    \caption{Illustration of several potential combinations of the prior adjacency matrix (\(A_{p1}\), \(A_{p2}\), \(A_{p3}\), \(A_{p4}\)). Each configuration integrates spatial (\(A_S\)), temporal (\(A_T\)), and spatio-temporal (\(A_{ST}\)) components across three time steps (\(t_1\), \(t_2\), \(t_3\)). For instance, \(A_{p1}\) focuses on temporal transitions (\(A_T\)) between specific pairs of consecutive time steps (e.g., \(t_1\)-\(t_2\) and \(t_2\)-\(t_1\)), while spatio-temporal relationships (\(A_{ST}\)) are emphasized for other consecutive pairs (e.g., \(t_2\)-\(t_3\) and \(t_3\)-\(t_2\)). By comparison, other configurations, such as \(A_{p4}\), provide a more uniform emphasis on spatio-temporal relationships across all time steps.}

    \label{fig:my_label}
\end{figure}

\paragraph{Task-specific Gating Mechanism}

We design a gating mechanism to obtain the final hybrid, task-specific adjacency matrices, denoted as \(\mathcal{A}^*_{k}\). Then, the obtained matrices will serve as the input into the respective task-specific GSTGC modules. This process employs a task-specific gating matrix \(M_{k}\) alongside the addition of the prior adjacency matrix \(A_P\) and a dynamic matrix \(B\). This can be mathematically represented as follows:

\begin{equation}
\mathcal{A}^*_{k} = M_k \odot (A_P + B) \in \mathbb{R}^{3N \times 3N}
\end{equation}

where \(\odot\) denotes element-wise multiplication, and the parameter \(M_K\) \(\in \mathbb{R}^{3N \times 3N}\) serves as a learnable matrix that modulates the contributions of both the prior spatio-temporal adjacency matrix \( A_P \) and the dynamic matrix \( B \). Incorporating task-specific gating matrices \(M_k\) contributes to the personalized construction of hybrid adjacency matrices according to the unique characteristics of each task, enhancing the framework's ability to capture and leverage the most relevant connections for each task. 

The hybrid adjacency matrix formulation helps to maintain stability in the HAMG module by balancing the contributions of the static prior matrix \( A_P \) and the dynamic matrix \( B \) . The static matrix provides a reliable structural foundation that anchors the learning process and mitigates the risk of instability when dealing with noisy or sparse data. Meanwhile, the dynamic matrix, learned from task data, introduces flexibility to capture evolving spatio-temporal relationships, enriching the representation of task-specific patterns. To ensure overall stability and avoid the introduction of additional noise, the task-specific gating mechanism selectively modulates the combination of \( A_P \) and \(B\). By keeping sparsity in the resulting hybrid adjacency matrix, the gating mechanism not only prevents overfitting to noisy connections but also ensures that only the most relevant and meaningful edges are retained. This design allows the adjacency matrix to balance structural integrity, task adaptability, and sparsity, supporting effective and robust learning across diverse tasks. 

\subsection{Group-wise Spatio-Temporal Graph Convolution Module}

The GSTGC module extends the capabilities of GCNs to better learn the complex spatio-temporal dependencies. The left side of Figure 3 illustrates the design of the GSTGC module. The key to this module is the use of group-wise spatio-temporal learning, with two different grouping operations that effectively distinguish between short-term interactions within three-time-step intervals and longer-term dependencies. Another important aspect of the GSTGC module is its incorporation of task-specific spatio-temporal adjacency matrices, generated by the HAMG module. 

\begin{figure*}[h!]
    \centering
    \includegraphics[width=0.7\textwidth, height=0.7\textheight, keepaspectratio]{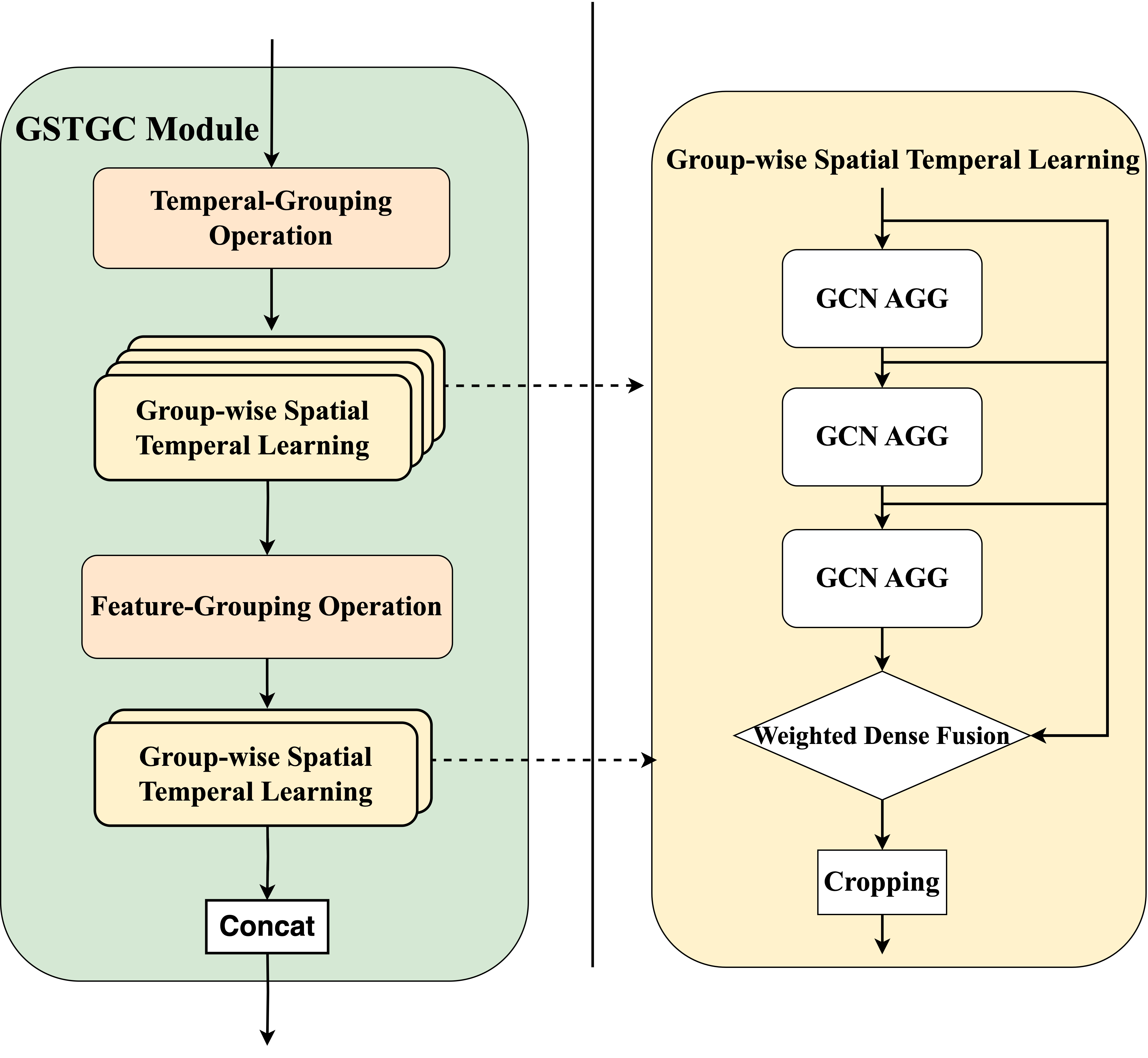}
    \caption{Illustration of the Group-wise Spatio-Temporal Graph Convolution Module}
    \label{fig:gcn}
\end{figure*}

In the GSTGC module, the first step is the temporal grouping operation of the input task-specific feature matrix \(X_k \). The temporal grouping is designed to isolate and model spatio-temporal dependencies within discrete time windows. In our methodology, the grouping operation specifically segments the feature matrix into sub-groups by partitioning the temporal dimension \(T\) into intervals of three consecutive time steps. This design choice is intentionally aligned with the dimensions of our adjacency matrix, which is structured as \(3N \times 3N\). This grouping operation could be formulated as:

\begin{equation}
Z = \mathcal{G_{T}}(X_k) = \left[ X_{k, :, 0:3, :}, X_{k, :, 3:6, :}, \ldots \right]
\end{equation}

where \(\mathcal{G}_T(\cdot)\) denotes the temporal grouping operation, \(X_k \) is the input feature matrix of the GSTGC module, where \( k \), ranging from 1 to \( K \), represents a specific prediction task. \( X_{k, :, 0:3, :}, X_{k, :, 3:6, :}, \ldots \) denotes the specific subgroups extracted from \(X_k \). The notation \(X_{k, :, i:j, :}
\) indicates a slice of \(X_k \) that includes all nodes, a specific range of time steps from \(i\) to \(j-1\), and all feature channels. This grouping strategy is designed with flexibility to suit various configurations. For instance, if the adjacency matrix is structured as \(2N \times 2N\) , the grouping operation would accordingly adjust to segment the input feature matrix across two consecutive time dimensions. Each grouped data is represented as\(\ Z_i \), which can then be used for further processing steps. Then, the grouped data is processed by a group-wise spatio-temporal learning blocks. In particular, group-wise spatio-temporal learning uses basic GCN aggregation (GCN-AGG) to learn spatio-temporal representations in the data. Each data group\(\ Z_i\) is subject to three consecutive GCN-AGG operations. This approach allows the independent processing of each group, which helps to identify and extract spatial and temporal features within short-term sequences. The mathematical formulation of the GCN-AGG operation is shown as follows:

\begin{equation}
H_i^{(g)} = \text{ReLU}\left(\mathcal{A}^*_{k} H_{i-1}^{(g)} W_i^{(g)} + \mathbf{b}_i^{(g)}\right) \in \mathbb{R}^{3N \times C'}
\end{equation}

where \(\mathcal{A}^*_{k}\) is the hybrid adjacency matrix for task \(k\) obtained in the HAMG module, \( H_{i-1}^{(g)} \) is the input into a GCN-AGG layer for group \( g \), \( W_i^{(g)} \in \mathbb{R}^{C \times C'} \) and \( \mathbf{b}_i^{(g)} \in \mathbb{R}^{C'} \) are learnable parameters, and \(\text{ReLU}\) is the non-linear activation function. For the first layer, \( H_0^{(g)} \) is set to the corresponding grouped data. Moreover, we introduce a weighted residual fusion module in our network to adaptively modulate the flow of information through the layers. This module integrates the outputs from each GCN-AGG layer along with the initial input \( H_0^{(g)} \), leveraging a set of learnable weights \( \mathbf{w}^{(g)} = [w_0^{(g)}, w_1^{(g)}, w_2^{(g)}, w_3^{(g)}] \). These weights are constrained to sum to 1, ensuring a normalized combination of information. Then, to ensure the dimensional coherence, the fused data is processed by a cropping operation to adjust the dimensions from \( 3N \) to \( N \) by selecting the middle segment \( N:2N \). This process can be formulated as follows:

\begin{equation}
    F_c^{(g)} = \mathcal{C}\left(\sum_{i=0}^{3} w_{i}^{(g)} H_i^{(g)}\right) \in \mathbb{R}^{N \times C'}
\end{equation}

where \( \mathcal{C}(\cdot) \)  denotes the cropping operation, \( F_{c}^{(g)} \) is the cropped output corresponding to group \(g\). In our framework, with a temporal dimension \(T\) = 12 and an adjacency matrix dimension of \(3N \times 3N\), we obtain four distinct outputs from the group-wise spatio-temporal learning blocks. The obtained four representations from the temporal grouping, along with the designed group-wise GCN module, no longer have a direct temporal dimension. Instead, they exist solely within the feature space. To address the need for capturing more complex interdependencies among features, we further introduce a feature grouping strategy. In this stage, overlapping combinations of the previously processed time blocks are formed (e.g., outputs 1-3 and outputs 2-4). The overlapping groups ensure temporal continuity, as adjacent time blocks can now jointly influence the next stages of processing. This could be formulated as follows:

\begin{equation}
F = \text{Concatenate}\left(F_c^{(1)}, F_c^{(2)}, F_c^{(3)}, F_c^{(4)}\right) \in \mathbb{R}^{N \times C' \times 4}
\end{equation}

\begin{equation}
M = \mathcal{G}_F(F) = \left[ F_{:,:,0:3}, F_{:,:,1:4} \right]
\end{equation}

where \( \mathcal{G}_F \) denotes the feature-grouping operation. Again, we use \( M_i \) to denote each grouped data. \( F_{:,:,0:3} \) represents the same slicing operation with the temporal grouping.

The use of an overlapping grouping strategy ensures that each feature group is not modeled in isolation but is contextualized within a broader framework. Then, similar procedure is implemented, employing one additional group-wise spatio-temporal learning block to further process each grouped feature matrix. By refining the learned feature representations, the model reflects subtle variations and intricate inter-feature relationships that evolve over time. The theoretical foundation for this feature grouping strategy is inspired by the concept of state space reconstruction in nonlinear time series analysis. In such analyses, a time series is reconstructed into a multidimensional state space to uncover the underlying dynamics of the system. In our model, each feature group created from overlapping time blocks acts as a reconstructed state, capturing the dynamic evolution of features over time. This approach mirrors how changes in one temporal block influence subsequent blocks, allowing the model to capture the intricate dependencies between different states effectively. This method—from temporal grouping to group-wise spatio-temporal learning, followed by feature grouping, and another learning block—enables the aggregation of information across multiple levels, thereby enhancing the model's ability to capture diverse spatio-temporal relationships. The final step involves the concatenation of the two resulting outputs followed by a max-pooling operation to distill the essential features from the aggregated data:

\begin{equation}
M_{\text{out}} = \text{MaxPool}\left(\text{Concatenate}\left(M_1, M_2\right)\right) \in \mathbb{R}^{N \times C'}
\end{equation}

\subsection{Other components of the framework}

\paragraph{Task-specific Input Layer}

Within our framework, each task's feature matrix initially passes through a task-specific input layer, constituted by a fully connected layer. This layer serves to project the input data into a higher-dimensional space, thereby enriching the data's representational capacity. The intent behind employing task-specific layers is to learn unique feature representations that are most relevant to each specific task. 

\paragraph{Multi-task Integration Unit}

We design a multi-task integration unit to combine information from various tasks for final predictions, effectively leveraging insights from each task to enhance overall prediction accuracy. For each task \(k\), the output obtained from the GSTGC module is represented by \( M_{\text{out}}^{k}\). These outputs are first concatenated to generate a unified matrix, containing information across all tasks. A task-agnostic output layer is then employed to obtain the final predictions, integrating two fully connected layers with a skip connection originating from the input merged feature matrix:

\begin{equation}
M' = \text{Concatenate}\left(M_{\text{out}}^1, M_{\text{out}}^2, \ldots, M_{\text{out}}^K\right) \in \mathbb{R}^{N \times C' \times K}
\end{equation}

\begin{equation}
Y_{\text{final}} = \text{ReLU}\left( W_1 M' + \mathbf{b}_1\right) W_2 + \mathbf{b}_2 + \mathbf{res} \in \mathbb{R}^{N \times K}
\end{equation}

where \( Y_{\text{final}} \) is the final predictions for all \(K\) tasks. The learnable parameters are defined as follows: \( W_1 \in \mathbb{R}^{C'K \times C_{\text{hidden}}}\), \( W_2 \in \mathbb{R}^{C_{\text{hidden}} \times K}\), \(\mathbf{b}_1 \in \mathbb{R}^{C_{\text{hidden}}}\), and \(\mathbf{b}_2 \in \mathbb{R}^{K}\). \(\mathbf{res}\) represents the skip connection from the input merged feature matrix, which is processed through a downsampling function to align with the dimensions.

\paragraph{Loss Function for Multi-Task Learning}

We employ a loss function that aggregates the loss from each individual task, applying task-specific scaling weights: 

\begin{equation}
Loss = \sum_{k=1}^{K} \beta_k  L_k
\end{equation}

where \(Loss\) is the total loss used for model training, \(\beta_1, \beta_2, \ldots, \beta_K\) are the scaling weights with the constraint that their sum equals 1. These weights ensure a balanced distribution of importance across all tasks. For individual task loss \(L_k\), we adopt the Smooth L1 loss \cite{huber1992robust} for its efficacy in regression tasks and its robustness to outliers. To tailor the loss magnitude to the specific requirements of different tasks, we introduce a pre-defined coefficient \(\alpha_k\) for controlling the magnitude of each task. The formulation of the Smooth L1 loss for a specific task \(k\) is given by:

\begin{equation}
L_k = 
\begin{cases}
0.5 \alpha_k (Y_k - \hat{Y}_k)^2 & \text{if } |Y_k - \hat{Y}_k| < \delta_k, \\
\alpha_k (|Y_k - \hat{Y}_k| - 0.5) & \text{otherwise}.
\end{cases}
\end{equation}

where \((Y_k - \hat{Y}_k)\) represents the difference between the true value \(Y_k\) and the predicted value \(\hat{Y}_k\), and \(\delta_k\) is the threshold that determines the switching behavior of the loss between the squared error and the absolute error.

\subsection{Complexity analysis}

Table 1 presents the complexity analysis of the proposed model's core components by module. We conducted both time and space complexity analyses for each module, focusing on the key contributors to the overall complexity. Specifically, the CTKE, HAMG, and GSTGC modules dominate the model’s computational load. The CTKE module includes operations like feature concatenation, embedding transformation, matrix multiplication, and softmax, while the HAMG module handles adjacency matrix operations. The GSTGC module is divided into temporal grouping and feature grouping components, each performing multi-layer GCNs. The overall model complexity is approximately the sum of these core modules, with minor contributions from smaller components like the input and output layers.

\begin{table}[htbp]
\centering
\caption{Time and Space Complexity of Each Module}
\label{tab:complexity_summary}
\footnotesize
\begin{tabular}{l p{5cm} p{5cm}}
\toprule
\textbf{Module} & \textbf{Time Complexity} & \textbf{Space Complexity} \\
\midrule
\textbf{CTKE Module} & $O(K \cdot N \cdot T \cdot C) + O(N \cdot C \cdot K \cdot D)$ \newline $+ O(N^2 \cdot D) + O(N^2)$ & $O(K \cdot N \cdot T \cdot C + N \cdot D + N^2)$ \\
\textbf{HAMG Module} & $O(N^2)$ & $O(N^2)$ \\
\textbf{GSTGC-1} & $O(N \cdot T \cdot C + T \cdot N^2 \cdot C$ \newline $+ N \cdot T \cdot C \cdot C')$ & $O(N \cdot T \cdot C + T \cdot N \cdot C $ \newline $+ N \cdot T \cdot C')$ \\
\textbf{GSTGC-2} & $O(N \cdot T \cdot C' + T \cdot N^2 \cdot C'$ \newline $+ N \cdot T \cdot C'^2)$ & $O(N \cdot T \cdot C'$ \newline $+ T \cdot N^2 \cdot C')$ \\
\bottomrule
\end{tabular}
\end{table}

\section{Experiments}

In this section, we conduct extensive experiments to compare our model with other state-of-the-art baseline methods for STMTL tasks. 

\subsection{Datasets}

Our framework is assessed using three real-world public datasets: the PEMSD4 and PEMSD8 datasets for traffic flow and speed predictions and the
NYCFHV dataset for forecasting pick-up and drop-off demand. The selection
of these distinct datasets aims to test our model’s ability to efficiently gen-
eralize across various types of MTL scenarios. Further details
on these datasets are provided in Table 2.

\paragraph{PEMSD4} The PEMSD4 dataset is a comprehensive highway traffic dataset collected by the Caltrans Performance Measurement System (PeMS). This dataset includes detailed information on both traffic flow and speed, gathered from 307 road detectors installed across 29 roads in the San Francisco Bay Area. The data collection period spans January and February of 2018. Traffic data is aggregated at 5-minute intervals, resulting in 12 data points per hour. The dataset is divided into training, validation, and test sets in a 6:2:2 ratio, providing a robust basis for evaluating model performance across different temporal segments. The geographic scope of the dataset encompasses the entire San Francisco Bay Area, offering a diverse and challenging environment for traffic prediction. The road detectors cover various types of roads, including highways and urban roads, capturing a wide range of traffic conditions and patterns. The high temporal resolution of the data ensures that the model can capture fine-grained temporal dynamics, which is crucial for accurate traffic prediction. The detailed traffic information provided by the PEMSD4 dataset makes it a popular benchmark for evaluating traffic forecasting models.

\paragraph{PEMSD8} The PEMSD8 dataset, like PEMSD4, is collected by the PeMS and provides traffic flow and speed data from 170 detectors in the San Bernardino area during the period of July to August 2016. While both datasets share similar temporal resolution and data aggregation structures, they exhibit distinct traffic patterns, as demonstrated in the Figure 4 below. These heatmaps represent the average flow and speed across all regions at the same time over two days. A clear difference in the spatial and temporal traffic dynamics can be observed between the two datasets, highlighting the unique characteristics of PEMSD8 compared to PEMSD4.

\paragraph{NYCFHV} The New York City For-Hire-Vehicle (NYCFHV) dataset is collected by the New York City Taxi and Limousine Commission. This dataset records ride-hailing activities by companies such as Uber and Lyft in New York City, covering over 200 million orders in 2022. It provides detailed information on pick-up and drop-off times and locations, allowing for precise spatio-temporal analysis of ride-hailing demand. Our study focuses on Manhattan, a densely populated area with a high demand for ride-hailing services. Manhattan is segmented into 69 specific regions. The orders are aggregated into 10-minute intervals, providing a high temporal resolution that is essential for accurate demand forecasting. The dataset is divided into training, validation, and test sets, with data from January to October 2022 used for training, November for validation, and December for testing.

\begin{table}[htbp]
\centering
\caption{Summary of Dataset Characteristics}
\label{tab:dataset_summary}
\footnotesize 
\begin{tabular}{lccc} 
\toprule
& \textbf{NYC dataset} & \textbf{PEMSD4 dataset} & \textbf{PEMSD8 dataset} \\ 
\midrule
Number of Nodes         & 69            & 307           & 170        \\ 
Number of Records       & 212,416,083   & 16,993        & 17,856     \\
Time Span               & 12 months     & 2 months      & 2 months   \\
Length of Time Interval & 10 min        & 5 min         & 5 min      \\
\bottomrule
\end{tabular}
\end{table}

\begin{figure*}[h!]
    \centering
    \includegraphics[width=1\textwidth, height=1\textheight, keepaspectratio]{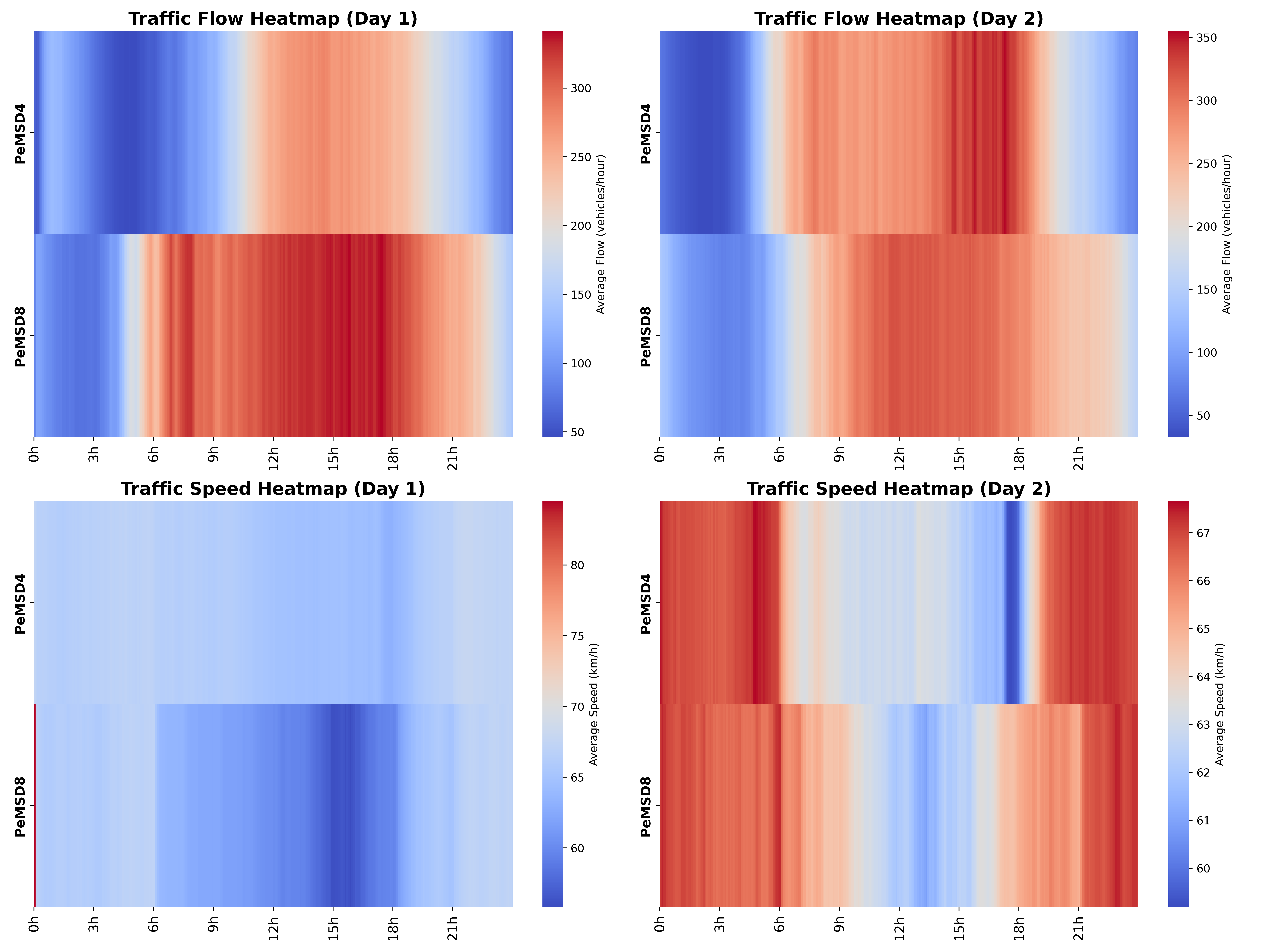}
    \caption{Heatmaps of average traffic flow and speed across all nodes for PEMSD4 and PEMSD8 over two days}
    \label{fig:gcn}
\end{figure*}

\subsection{Evaluation Metrics}

To evaluate the predictive accuracy of our model on the datasets, we employ three error metrics: Mean Absolute Error (MAE), Root Mean Squared Error (RMSE), and Mean Average Percentage Error (MAPE):

\begin{equation}
    \text{MSE} = \frac{1}{n} \sum_{i=1}^{n} (Y_i - \hat{Y}_i)^2
\end{equation}

\begin{equation}
    \text{RMSE} = \sqrt{\frac{1}{n} \sum_{i=1}^{n} (Y_i - \hat{Y}_i)^2}
\end{equation}

\begin{equation}
    \text{MAE} = \frac{1}{n} \sum_{i=1}^{n} |Y_i - \hat{Y}_i|
\end{equation}

where \( n \) represents the total number of observations, \( Y_i \) is the actual value of the \( i^{th} \) observation, and \( \hat{Y}_i \) is the predicted value.

\subsection{Baselines Methods and Experiment Settings}

Following previous studies \cite{zhang2023autostl}, we predict the traffic attribute for the next time interval using the past 12 time steps. In our study, we compare the performance of our proposed model against several baselines include Long short-term memory (LSTM), Random Forest (RF), XGBoost, Gradient-Boosted Decision Trees (GBDT), Multi-Layer Perceptron (MLP), Temporal Graph Convolutional Network (T-GCN) \cite{zhao2019t}, Spatial-Temporal Graph Convolutional Network (STGCN) \cite{yu2017spatio}, Spatial-Temporal Synchroous Graph Convolutional Network (STSGCN) \cite{song2020spatial}, Multi-View Spatial-Temporal Graph Convolutional Network (MSTGCN) \cite{guo2019attention} and and Spatio-Temporal Adaptive Embedding transformer (STAEformer) \cite{liu2023spatio}. For traditional machine learning and deep learning models, hyperparameter tuning is manually conducted over a comprehensive parameter space. For other models, we employ the configurations as specified in their respective original publications, utilizing the PyTorch Geometric Temporal library.  For our model, we perform a comprehensive hyperparameter analysis to investigate the effects of various configuration settings. Detailed results of these experiments are presented in Section 5.6. We use the Adam optimization combined with an early stopping mechanism, applying a patience of 10 epochs. The maximum epoch is set to 200. The implementation of neural network-based models is carried out using the PyTorch framework, with computations performed on cloud-based high performance computing infrastructure provided by King's College London. The average training time per epoch is approximately 1414 seconds for the PEMSD4 dataset and 2478 seconds for the NYC dataset. During inference, the model achieves an average inference time of 13.60 milliseconds for PEMSD4 and 50.20 milliseconds for NYC on the devices utilized in our study.

\subsection{Experimental Results and Analysis}

Tables 3, 4, 5 and 6 present the performance metrics of different models in three datasets. Our model shows consistently improved performance across all MTL tasks, indicating its effectiveness and generalization in capturing complex spatio-temporal patterns and MTL capability. Below we detail the experimental results and provide a comparative discussion.

\paragraph{Result on PEMSD4 dataset} 

In Table 3, it can be seen that our model outperforms all traditional machine learning and GCN-based methods. Specifically, the proposed model achieves substantial reductions in RMSE, MAE, and MAPE by 89.33\%, 92.01\%, and 92.43\% for the speed task, and 56.33\%, 59.28\%, and 61.77\% for the flow task, compared to the best performance of traditional models. These improvements are significant as they indicate a much higher accuracy in predictions, which is crucial for applications such as traffic management and urban planning. When compared to the best graph-based model, our proposed model also demonstrates notable improvements in the speed prediction task. Specifically, we observed a reduction in the RMSE by 6.31\%, MAE by 19.32\%, and MAPE by 16.27\%. In the context of flow prediction task, the improvements are even more significant. The reductions are 42.27\% for RMSE, 54.96\% for MAE, and 99.50\% for MAPE. Compared with STAEformer, the current state-of-the-art model employing multiple learnable embedding matrices and a transformer structure, our proposed model still demonstrates competitive performance. Specifically, for the speed prediction task, our model achieves a better RMSE, although the MAE is slightly higher. For the flow prediction task, similar to the comparison with the best-performing graph-based model, our approach significantly outperforms STAEformer, achieving improvements of 29.42\% in RMSE, 42.94\% in MAE, and 37.74\% in MAPE.

\paragraph{Result on PEMSD8 dataset}

Result on PEMSD8 dataset. In Table 5, it is clear that our proposed model outperforms all traditional machine learning methods, achieving the best performance across all three metrics (RMSE, MAE, and MAPE) for both speed and flow prediction tasks. Compared to the best-performing traditional model (RF), our model delivers significantly more accurate predictions, demonstrating its ability to handle the complex spatio-temporal dependencies inherent in traffic data. Furthermore, our model also outperforms the best graph-based model (MSTGCN), achieving better results in all three metrics for both tasks. This indicates that our proposed approach captures richer spatio-temporal relationships, enabling more precise predictions. The multi-task framework utilized in our model further enhances its capability to jointly learn speed and flow dynamics, providing a more holistic understanding of traffic patterns and improving overall prediction accuracy. Compared with STAEformer, our proposed model remains competitive on this dataset. Specifically, for speed prediction, our model performs slightly worse in RMSE (1.623 vs. 1.602) and MAE (0.815 vs. 0.799), while achieving equal performance in MAPE. However, for flow prediction, our model again shows substantial improvements, reducing RMSE by 31.11\%, MAE by 41.97\%, and MAPE by 35.06\%.

\begin{table}[htbp]
\caption{Performance Comparison of Models on the PEMSD4 Dataset for Speed and Flow Prediction}
\label{tab:traffic_prediction}
\centering
\resizebox{\columnwidth}{!}{%
\begin{tabular}{lcccccc}

\midrule
& \multicolumn{3}{c}{Speed} & \multicolumn{3}{c}{Flow} \\
\cmidrule(lr){2-4} \cmidrule(lr){5-7}
Model & RMSE & MAE & MAPE & RMSE & MAE & MAPE \\
\midrule
LSTM & 25.446 & 12.539 & 0.235 & 59.180 & 29.533 & 0.190 \\
RF & 16.766 & 12.386 & 0.243 & 74.124 & 54.155 & 0.254 \\
XGBoost & 20.363 & 14.482 & 0.382 & 72.522 & 52.888 & 0.253 \\
GBDT & 19.568 & 13.460 & 0.348 & 63.260 & 45.715 & 0.218 \\
MLP & 17.032 & 6.529 & 0.117 & 43.806 & 23.405 & 0.181 \\
T-GCN & 1.909 & 1.227 & 0.022 & 37.260 & 23.220 & 15.590 \\
STGCN & 3.402 & 2.043 & 0.040 & 34.890 & 21.160 & 13.830 \\
STSGCN & 7.916 & 4.334 & 0.106 & 33.720 & 21.430 & 14.830 \\
MSTGCN & 2.904 & 1.905 & 0.037 & 37.210 & 23.960 & 14.330 \\
STAEformer & 1.796 & \textbf{0.925} & \textbf{0.018} & 27.103 & 16.704 & 0.111 \\
\textbf{Proposed model} & \textbf{1.789}$^\dagger$ & 0.990$^\dagger$ & \textbf{0.018}$^\dagger$ & \textbf{19.129}$^\dagger$ & \textbf{9.531}$^\dagger$ & \textbf{0.069}$^\dagger$ \\
\bottomrule
\end{tabular}%
}
\begin{tablenotes}
\small
\item $^\dagger$: Results are statistically significant with $p < 0.01$.
\end{tablenotes}
\end{table}

\begin{table}[!h]
\caption{Performance Comparison of Spatio-Temporal Multi-Task Learning Models on the PEMSD4 Dataset (adapted from \cite{zhang2023autostl})}
\label{tab:model_comparison}
\centering
\begin{tabular}{lcccc}
\toprule
\multicolumn{1}{c}{Model} & \multicolumn{2}{c}{Speed} & \multicolumn{2}{c}{Flow} \\
\cmidrule(lr){2-3} \cmidrule(lr){4-5}
 & RMSE & MAE & RMSE & MAE \\
\midrule
CCRNN & 2.560 & 1.460 & 27.990 & 17.590 \\
DCRNN & 2.340 & 1.520 & 27.650 & 17.440 \\
MTGNN & 2.050 & 1.120 & 28.030 & 17.920 \\
AutoSTL & 2.010 & 1.080 & 27.360 & 17.400 \\
PLE & 2.250 & 1.290 & 30.140 & 19.240 \\
\textbf{Proposed model} & \textbf{1.789} & \textbf{0.990} & \textbf{19.129} & \textbf{9.531} \\
\bottomrule
\end{tabular}
\end{table}

\begin{table}[htbp]
\caption{Performance Comparison of Models on the PEMSD8 Dataset for Speed and Flow Prediction}
\label{tab:traffic_prediction_pemsd8}
\centering
\resizebox{\columnwidth}{!}{%
\begin{tabular}{lcccccc}
\toprule
& \multicolumn{3}{c}{Speed} & \multicolumn{3}{c}{Flow} \\
\cmidrule(lr){2-4} \cmidrule(lr){5-7}
Model & RMSE & MAE & MAPE & RMSE & MAE & MAPE \\
\midrule
LSTM & 6.424 & 3.771 & 0.082 & 42.902 & 29.173 & 0.191 \\
RF & 4.673 & 2.905 & 0.066 & 35.476 & 30.499 & 0.186 \\
XGBoost & 8.436 & 8.277 & 0.145 & 42.409 & 32.695 & 0.193 \\
GBDT & 5.576 & 3.055 & 0.069 & 45.811 & 36.088 & 0.195 \\
MLP & 10.741 & 10.164 & 0.170 & 44.917 & 32.429 & 0.220 \\
T-GCN & 2.182 & 1.184 & 0.021 & 35.790 & 22.722 & 0.140 \\
STGCN & 2.775 & 1.788 & 0.033 & 27.091 & 17.501 & 0.113 \\
STSGCN & 4.093 & 2.008 & 0.053 & 26.803 & 17.134 & 0.110 \\
MSTGCN & 2.139 & 1.475 & 0.025 & 29.151 & 19.001 & 0.124 \\
STAEformer & \textbf{1.602} & \textbf{0.799} & \textbf{0.015} & 19.626 & 11.776 & 0.077 \\
\textbf{Proposed model} & 1.623$^\dagger$ & 0.815$^\dagger$ & \textbf{0.015}$^\dagger$ & \textbf{13.521}$^\dagger$ & \textbf{6.834}$^\dagger$ & \textbf{0.050}$^\dagger$ \\
\bottomrule
\end{tabular}%
}
\begin{tablenotes}
\small
\item  $^\dagger$: Results are statistically significant with $p < 0.01$.
\end{tablenotes}
\end{table}

\paragraph{Result on NYCFHV dataset} Similar to the results from the PEMSD4 dataset, our model significantly outperforms both traditional methods, GCN-based prediction models, and state-of-the-art transformer-based model in nearly all performance metrics. Specifically, for pick-up demand, our model demonstrates reductions of 25.64\% in RMSE, 17.82\% in MAE, and 13.07\% in MAPE. For drop-off demand, the reductions are even more significant: 29.01\% in RMSE, 21.27\% in MAE, and 26.00\% in MAPE. Compared to GCN-based baseline models, including T-GCN, STGCN, STSGCN, and MSTGCN, our model exhibits considerable improvements, with reductions of 10.66\% in RMSE and 5.79\% in MAE for pick-up demand, and 5.94\% in RMSE and 0.55\% in MAE for drop-off demand. In terms of MAPE, our model achieves the second-best performance for both pick-up and drop-off demand tasks, slightly performing behind the T-GCN for pick-up demand and STSGCN for drop-off demand. Regarding MAPE, our model ranks third for both pick-up and drop-off demand predictions, showing slightly lower performance compared to T-GCN and STAEformer on pick-up demand, and STSGCN and STAEformer on drop-off demand. Nonetheless, it still maintains competitive predictive accuracy overall.

\begin{table}[!h]
\caption{Performance Comparison of Models on NYCFHV Dataset for Pick-up and Drop-off Demand Prediction}
\label{tab:performance_nycfhv}
\centering
\resizebox{\columnwidth}{!}{%
\begin{tabular}{lcccccc}
\toprule
\multicolumn{1}{c}{} & \multicolumn{3}{c}{Pick-up Demand} & \multicolumn{3}{c}{Drop-off Demand} \\
\cmidrule(lr){2-4} \cmidrule(lr){5-7}
Model & RMSE & MAE & MAPE & RMSE & MAE & MAPE \\
\midrule
LSTM & 8.332 & 5.478 & 0.329 & 10.784 & 6.715 & 0.417 \\
RF & 9.080 & 6.113 & 0.345 & 17.529 & 13.837 & 0.600 \\
XGBoost & 8.971 & 5.942 & 0.345 & 17.563 & 13.825 & 0.600 \\
GBDT & 8.289 & 5.621 & 0.333 & 17.496 & 13.795 & 0.596 \\
MLP & 9.591 & 5.782 & 0.344 & 8.196 & 5.437 & 0.395 \\
T-GCN & 7.404 & 5.041 & 0.281 & 6.538 & 4.576 & 0.293 \\
STGCN & 6.935 & 4.778 & 0.299 & 7.792 & 4.847 & 0.341 \\
STSGCN & 6.995 & 4.786 & 0.285 & 6.178 & 4.304 & 0.281 \\
MSTGCN & 7.257 & 4.992 & 0.299 & 6.771 & 4.671 & 0.317 \\
STAEformer & 6.931 & 4.701 & \textbf{0.262} & 6.274 & 4.346 & \textbf{0.275} \\
\textbf{Proposed model} & \textbf{6.196}$^\dagger$ & \textbf{4.501}$^\dagger$ & 0.286$^\dagger$ & \textbf{5.811}$^\dagger$ & \textbf{4.280}$^\dagger$ & 0.293$^\dagger$ \\
\bottomrule
\end{tabular}%
}
\begin{tablenotes}
\small
\item  $^\dagger$: Results are statistically significant with $p < 0.01$.
\end{tablenotes}
\end{table}

\paragraph{Comparison with other spatio-temporal multi-task frameworks}

To provide a thorough comparative analysis, we benchmarked our model against state-of-the-art GCN-based models specifically designed for STMTL. The results of this comparison are detailed in Table 4. It is worth mentioning that models specifically designed for multi-task prediction show similar performance levels. However, our model significantly outperforms the best of these models in terms of RMSE and MAE. Specifically, for flow predictions, our model reduces RMSE and MAE by 30.09\% and 45.23\%, respectively. For speed predictions, the reductions are 11.01\% in RMSE and 8.38\% in MAE. This indicates our model's superior ability to capture and predict the complex dynamics of spatio-temporal data.

\paragraph{Overall Performance Analysis} 

Our analysis covers a wide range of models applied to different multi-task spatio-temporal scenarios, particularly focusing on the traffic speed and flow prediction on the PEMSD4 and PEMSD8 datasets, as well as ride-hailing pick-up and drop-off demand prediction on the NYCFHV dataset. Traditional machine learning models exhibit limitations in their performance. This shortcoming is primarily attributed to their inherent limitations in capturing the complex interconnections between spatial and temporal dependencies, which is crucial for accurately forecasting dynamics. Neural network-based models generally perform better than traditional machine learning models in many prediction tasks. However, when it comes to multi-task spatio-temporal settings, their effectiveness is not always consistent. Firstly, while LSTM is widely recognized for its strength in modeling time-series data, it performs at a level similar to or even below that of MLP. One key reason is that both LSTM and MLP are primarily designed to handle Euclidean data. In reality, traffic networks are complicated, with varying distances and connections between nodes (such as highways and urban roads) that cannot be mapped onto a grid. Therefore, LSTM and MLP struggle to effectively leverage the rich interaction information within these networks. Moreover, both neural network and machine learning models tend to model the temporal aspects of data without sufficiently considering dependencies between nodes. In contrast, graph-based models like TGCN, STGCN, STSGCN, MSTGCN and ours, specifically designed for spatio-temporal predictions, show superior performance. This enhanced capability is due to their adeptness at processing graph-structured, non-Euclidean data, which is critical for capturing the dynamics of complex networks. Previous graph-based models, while outperforming other traditional baselines in modeling spatio-temporal relationships, often have challenges in effectively handling multiple prediction tasks simultaneously. Our proposed model mitigates these shortcomings by introducing a novel STMTL framework. The adjacency matrix generation module contributes to the enhanced performance of our model. The enhancements of this module are twofold. Firstly, by incorporating both a dynamic adjacency matrix, which learns from input spatio-temporal data, and a static one, predefined based on prior knowledge, our adjacency matrix effectively captures the essence of real-world networks as well as the evolving dynamics of different regimes. This dual approach ensures that the representation is both accurate and adaptable to changes in the environment. Secondly, through the introduction of a gating mechanism specifically designed for MTL, we ensure that the adjacency matrix is task-specific, representing the unique characteristics of each task while facilitating the sharing of crucial information across tasks. This not only enhances task-specific performance but also leverages the synergistic benefits inherent in MTL. Moreover, the architecture of the GSTGC module also improves the modeling capability of complex spatio-temporal dependencies, contributing to the state-of-the-art performance.

Notably, we also compared our model with the current state-of-the-art model—STAEformer, which utilizes transformer-based architectures rather than graph-based methods. Across all three evaluated datasets, our model consistently demonstrates more robust overall performance. Particularly in flow prediction tasks, our model achieves notable margins of improvement over STAEformer. This significant advantage may stem from our model's capability to effectively leverage cross-task information.Such synergy between tasks allows our model to better represent complex dependencies and interactions, ultimately resulting in superior predictive performance.

\subsection{Ablation study}

To systematically study the contributions of key components within our model, we conduct ablation studies across both datasets.

\paragraph{Effect of the information sharing mechanism } Firstly, we focus on evaluating the role of several information sharing mechanisms employed in our framework. Accordingly, we examine the model under three specific configurations: \textbf{Variant1} examines the model's performance without the CTKE unit, relying solely on a static adjacency matrix without any dynamically generated components. \textbf{Variant2} examines the model's performance by relying solely on the dynamically learned adjacency matrix, removing the static component. \textbf{Variant3} investigates the model's performance without the task-specific gating matrix \(M_k\), focusing solely on the combined influence of static and dynamic adjacency matrices without task-specific modulation. \textbf{Variant4} assesses the model's performance without the task-agnostic output layer, examining how the absence of this component affects the integration and final prediction across multiple tasks.  Variant 5 assesses the model's performance without the task-specific input layer. 

As shown in Table 7, Variant1, which utilizes a static predefined adjacency matrix, exhibits a notable decrease in performance compared to the full model. This shows the limitations of static matrices in capturing the complex dependencies inherent in spatio-temporal data. This advantage stems from the inherent flexibility of dynamically learnable adjacency matrices, which continuously refine their structure to align with the emerging patterns observed in the dataset. Variant 2, which relies solely on the dynamically learned adjacency matrix B and removes the static prior static also demonstrates a significant performance drop. While the dynamic adjacency matrix can adapt to evolving patterns, its lack of prior structural information leads to instability and overfitting, particularly when handling noisy or sparse data. This result highlights the importance of incorporating a stable prior to ensure robust performance. The observed performance drop in Variant3, which lacks the task-specific gating matrix \(M_k\), highlights the essential function of this matrix in our framework. By modulating the contributions of both the static and dynamic adjacency matrices, the gating matrix adjusts the influence of both static and dynamic adjacency matrices, ensuring that each task receives the most relevant spatio-temporal information. This approach not only allows the model to effectively leverage the benefits of both static and dynamic adjacency matrices but also supports task-focused information flow. Taken together, the ablation results highlight the necessity of combining static and dynamic adjacency matrices with the task-specific gating mechanism. Relying solely on static or dynamic adjacency matrices leads to performance degradation—static matrices lack adaptability to evolving patterns, while dynamic matrices, without prior structural information, are prone to instability and overfitting. Moreover, a simple combination of the two, without gating, fails to effectively balance their contributions across tasks. Our proposed approach addresses these limitations by leveraging the complementary strengths of static and dynamic components, while the gating mechanism dynamically adjusts their contributions based on task-specific requirements. This design not only ensures robustness through the incorporation of prior structural knowledge but also enables flexibility to capture emerging spatio-temporal dependencies, thereby achieving superior performance and generalization. Variant4 of our study, which omits the task-agnostic output layer, also shows decreased performance compared to the original settings, highlighting the layer's crucial role in effectively integrating and coordinating outputs across multiple tasks. This finding substantiates the importance of a unified output layer in enhancing the model’s ability to generalize across diverse learning tasks. Variant 5, which replaces the task-specific input layer with a shared input layer across all tasks, also demonstrates a noticeable performance drop. While a shared input layer simplifies the parameter structure and promotes shared representations, it fails to capture task-specific nuances effectively. This lack of task differentiation dilutes the model's ability to adapt to the unique characteristics of each task, leading to suboptimal performance. This result underscores the critical role of task-specific input layers in preserving task-relevant information and enabling more effective MTL.

To further investigate the convergence properties and stability of the HAMG module, we compared the training loss curves for Variant 1 (static-only adjacency matrix), Variant 2 (dynamic-only adjacency matrix), Variant 3 (hybrid adjacency matrix without the gating mechanism), and the full Model (hybrid adjacency matrix with the gating mechanism). As shown in Figure 5, all configurations exhibit smooth and stable convergence during training, indicating that the proposed HAMG module maintains robust learning dynamics across different adjacency matrix settings. Notably, the Full Model achieves the best overall performance while maintaining comparable convergence stability. This demonstrates that the task-specific gating mechanism not only enhances the model's ability to adaptively balance the contributions of static and dynamic components, but also introduces no adverse effects on the training process. The results collectively validate the stability and practicality of our design choices in constructing the hybrid adjacency matrix.

\begin{figure*}[h!]
    \centering
    \includegraphics[width=1\textwidth, height=1\textheight, keepaspectratio]{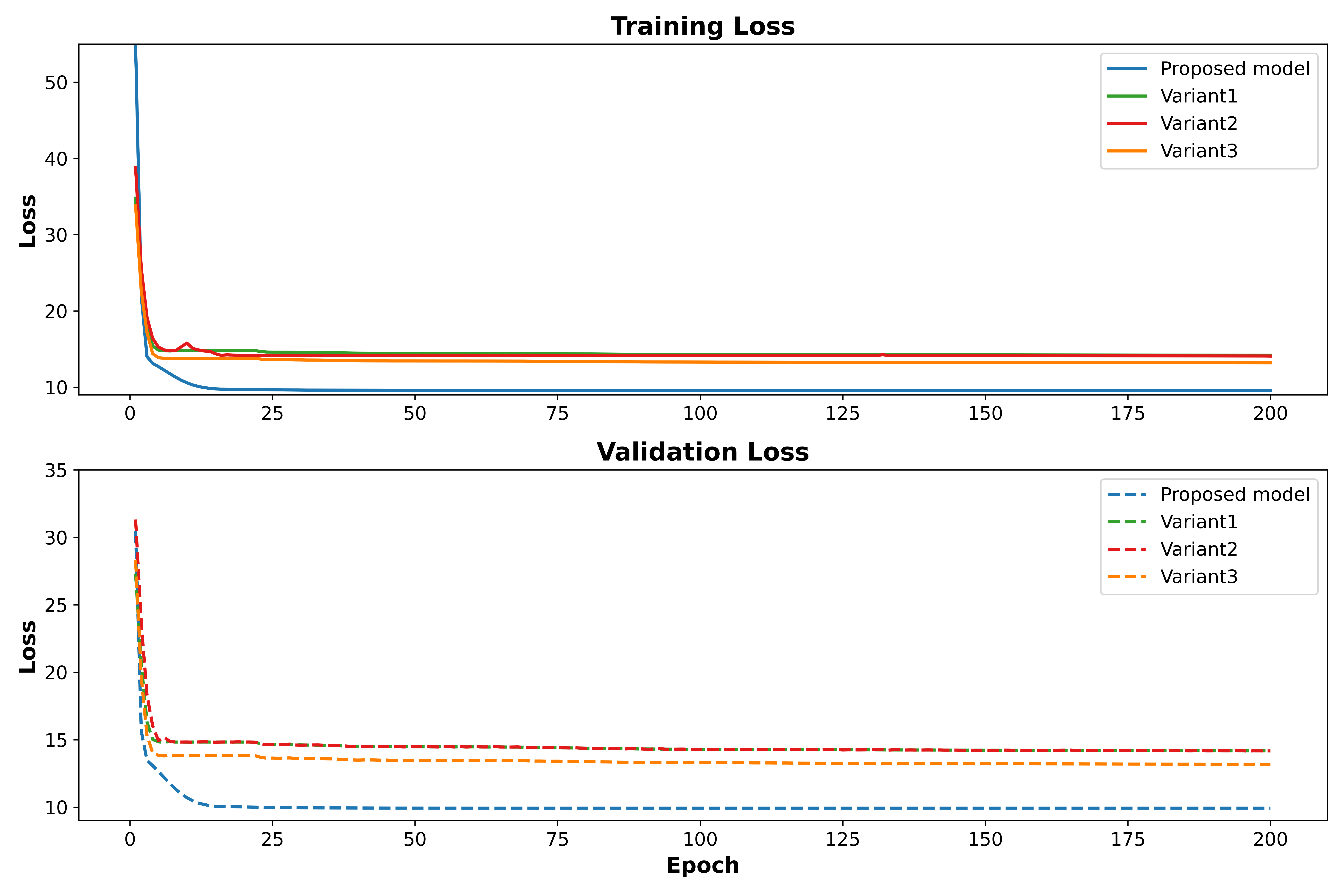}
    \caption{Comparison of Training and Validation Loss Across Variant1, Variant2, Variant3 and Proposed model}
    \label{fig:gcn}
\end{figure*}

\paragraph{Effect of Prior spatio-temporal Adjacency Matrix} We investigate the impact of different prior spatio-temporal adjacency matrix combinations on our model's performance. Figure 2 illustrates several potential combinations, labeled as \(\mathcal{A}_{P2}\), \(\mathcal{A}_{P3}\), and \(\mathcal{A}_{P4}\), in contrast to \(\mathcal{A}_{P1}\) employed by our original model. We test the following variants: \textbf{Variant 6} employs \(\mathcal{A}_{P2}\) for its prior adjacency matrix. \textbf{Variant 7} employs \(\mathcal{A}_{P3}\). \textbf{Variant 8} employs \(\mathcal{A}_{P4}\). While these configurations slightly influence the prediction outcomes, the overall impact remains relatively modest. Past research has explored the use of search algorithms to identify the optimal configuration of spatio-temporal adjacency matrices. Such approaches have demonstrated potential improvements in model performance by dynamically adapting the adjacency matrix to better reflect the underlying dynamics. Future studies could consider employing more advanced methods, such as reinforcement learning, for the dynamic selection of optimal adjacency matrix configurations.

\paragraph{Effect of weighted dense residual connections} Finally, we focus on the components within the GSTGC module. Specifically, we experiment the following variant: \textbf{Variant 9} investigates the effect of the dense residual connection within the group-wise spatio-temporal learning block. The results of Variant7 demonstrate a notable decrease in performance metrics across all evaluated tasks when compared to the proposed model. The observed decline implies the significance of dense residual connections in mitigating potential issues associated with GCN convolutions, such as oversmoothing. Dense residual connections help to preserve the richness of the feature space by allowing information from earlier layers to be directly carried forward, thus maintaining distinctiveness among node features and enhancing the model's ability to capture complex spatio-temporal relationships.

\paragraph{Effect of two group operations} To validate the importance of the group operations, we conducted experiments using two model variants. In \textbf{Variant 10}, we replaced the temporal grouping operation with a simple MLP, while in \textbf{Variant 11}, both the feature grouping and the GCN module were replaced by a simple MLP. The results on two datasets consistently showed that removing either group operation led to a significant decline in performance, indicating the effectiveness of these operations. The temporal grouping operation allows the model to learn more localized spatio-temporal representations by grouping data along the temporal dimension, enabling the GCN module to capture temporal dependencies effectively. On the other hand, feature grouping enhances performance by refining the learned representations from each time step, allowing the model to better capture complex spatio-temporal relationships. By reorganizing features after initial temporal processing, the model can integrate and process these features more deeply.

\begin{table*}[htbp]
\caption{Comprehensive Performance Comparison of Various Models on the PEMSD4 and NYC Datasets for Speed, Flow, Pick-up Demand, and Drop-off Demand Prediction}
\label{tab:comprehensive_performance_comparison}
\centering
\resizebox{\textwidth}{!}{%
\begin{tabular}{@{}lcccccccccccc@{}}
\toprule
& \multicolumn{6}{c}{PEMSD4 dataset} & \multicolumn{6}{c}{NYC dataset} \\
\cmidrule(lr){2-7} \cmidrule(lr){8-13}
& \multicolumn{3}{c}{Speed} & \multicolumn{3}{c}{Flow} & \multicolumn{3}{c}{Pick-up Demand} & \multicolumn{3}{c}{Drop-off Demand} \\
\cmidrule(lr){2-4} \cmidrule(lr){5-7} \cmidrule(lr){8-10} \cmidrule(lr){11-13}
Model & RMSE & MAE & MAPE & RMSE & MAE & MAPE & RMSE & MAE & MAPE & RMSE & MAE & MAPE \\
\midrule
Variant1                                & 2.019         & 1.131         & 0.022     & 20.331        & 10.155        & 0.074     & 6.338         & 4.607                    & 0.294            & 5.926         & 4.349         & 0.296     \\
Variant2                                       & 2.426         & 1.248     & 0.025       & 19.932         & 10.031     & 0.074        & 6.565 & 4.731	& 0.302	       &  6.012	       & 4.542	       & 0.311       \\
Variant3                                & 1.829         & 1.006         & 0.019     & 19.669        & 9.824         & 0.070     & 6.387         & 4.648                    & 0.298            & 6.066         & 4.465         & 0.307     \\
Variant4                                & 1.820         & 1.001         & 0.019     & 19.671        & 9.819         & 0.070     & 6.414         & 4.669                    & 0.298            & 6.105         & 4.496         & 0.310     \\
Variant5                                & 2.446	       & 1.258	        & 0.025	     & 19.935	       & 10.037	        & 0.074	    & 6.549	      & 4.673	                        & 0.299	             & 6.000	         & 4.535	        & 0.307     \\
Variant6                                & 1.792         & 0.986         & 0.018     & 19.306        & 9.633         & 0.070     & 6.346         & 4.608                    & 0.293            & 5.938         & 4.356         & 0.296     \\

Variant7                                & 1.797         & 0.990         & 0.018     & 19.454        & 9.693         & 0.070     & 6.340         & 4.607                    & 0.294            & 5.927         & 4.354         & 0.297     \\
Variant8                                & 1.794         & 0.990         & 0.018     & 19.425        & 9.680         & 0.070     & 6.303         & \textbf{4.433}           & \textbf{0.281}   & 5.857         & 4.243         & \textbf{0.290} \\
Variant9                                & 1.796         & 0.990         & 0.018     & 19.455        & 9.693         & 0.070     & 6.338         & 4.606                    & 0.294            & 5.924         & 4.352         & 0.298     \\
Variant10                                & 1.968         & 1.050         & 0.020     & 19.782        & 9.836         & 0.071     & 6.362         & 4.637                    & 0.300            & 5.958         & 4.395         & 0.307     \\
Variant11                                & 1.969         & 1.050         & 0.020     & 19.777        & 9.835         & 0.071     & 6.351         & 4.628                    & 0.302            & 5.946         & 4.377         & 0.300     \\
\textbf{Proposed model}                 & \textbf{1.789} & \textbf{0.990} & \textbf{0.018} & \textbf{19.129} & \textbf{9.531} & \textbf{0.069} & \textbf{6.196} & 4.501 & 0.286 & \textbf{5.811} & \textbf{4.280} & 0.293 \\
\bottomrule
\end{tabular}%
}
\end{table*}

\subsection{Hyperparameter analysis}

In our study, we conducted detailed hyperparameter experiments to evaluate the impact of various configurations on the predictive performance of our model using the PEMSD4 dataset. We focus on batch size, learning rate, GCN hidden layer dimensions, and the number of GCN layers.  Figure 6 shows the results of these experiments in terms of RMSE, MAE, and MAPE. For batch size, the results indicat that smaller batch sizes generally yield better results, with an optimal performance observed at a batch size of 24 before a decline as batch size increases. When exploring the number of layers, performance significantly improves moving from one to three layers, suggesting that a deeper network is more capable of capturing complex patterns; however, adding a fourth layer offers no substantial improvement, indicating a plateau in benefits with additional depth. For learning rates, the model shows optimal performance at a rate of 0.003, balancing fast convergence without the instability seen at higher rates. For hidden dimensions, the results indicate that increasing the dimension up to 64 improves the model's performance. However, further increases lead to a significant collapse in performance, suggesting potential overfitting issues.

\begin{figure}[H]
\centering
\includegraphics[width=\textwidth, height=1\textheight, keepaspectratio]{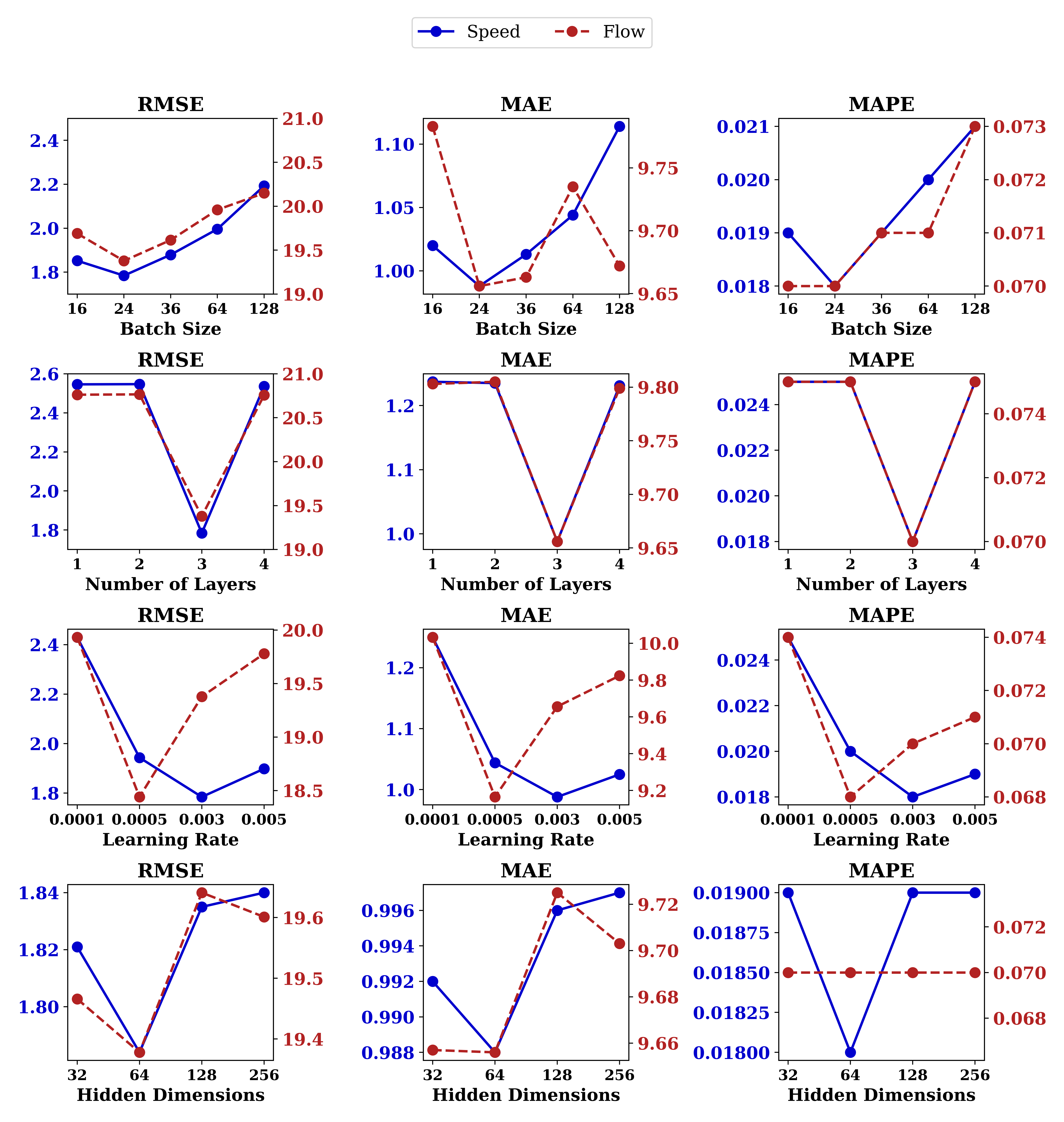}
\caption{Influence of hyperparameters on PEMSD4 dataset}
\label{fig:image2}
\end{figure}

\subsection{Model interpretation}

To better understand the proposed model, we select node no. 60 with flow and speed data from the PeMSD4 dataset and visualize the results of the test set. We generate predictions for this node using our model and compared these with the actual values for epochs 5, 15, and 30. Figure 7 shows noticeable improvements as training progresses. At epoch 5, the model’s predictions are roughly aligned with the ground truth values but with significant errors, especially during high and low peaks. By epoch 15, the model better follows the trends, though it still tends to miss during rapid changes in flow and speed. It is only by epoch 50 that the predictions closely match the actual values, indicating a significant increase in the model's prediction accuracy. These observations suggest a positive correlation between the number of training iterations and the model's ability to generalize from the data.

\begin{figure}[H]
\centering
\includegraphics[width=\textwidth, height=1.2\textheight, keepaspectratio]{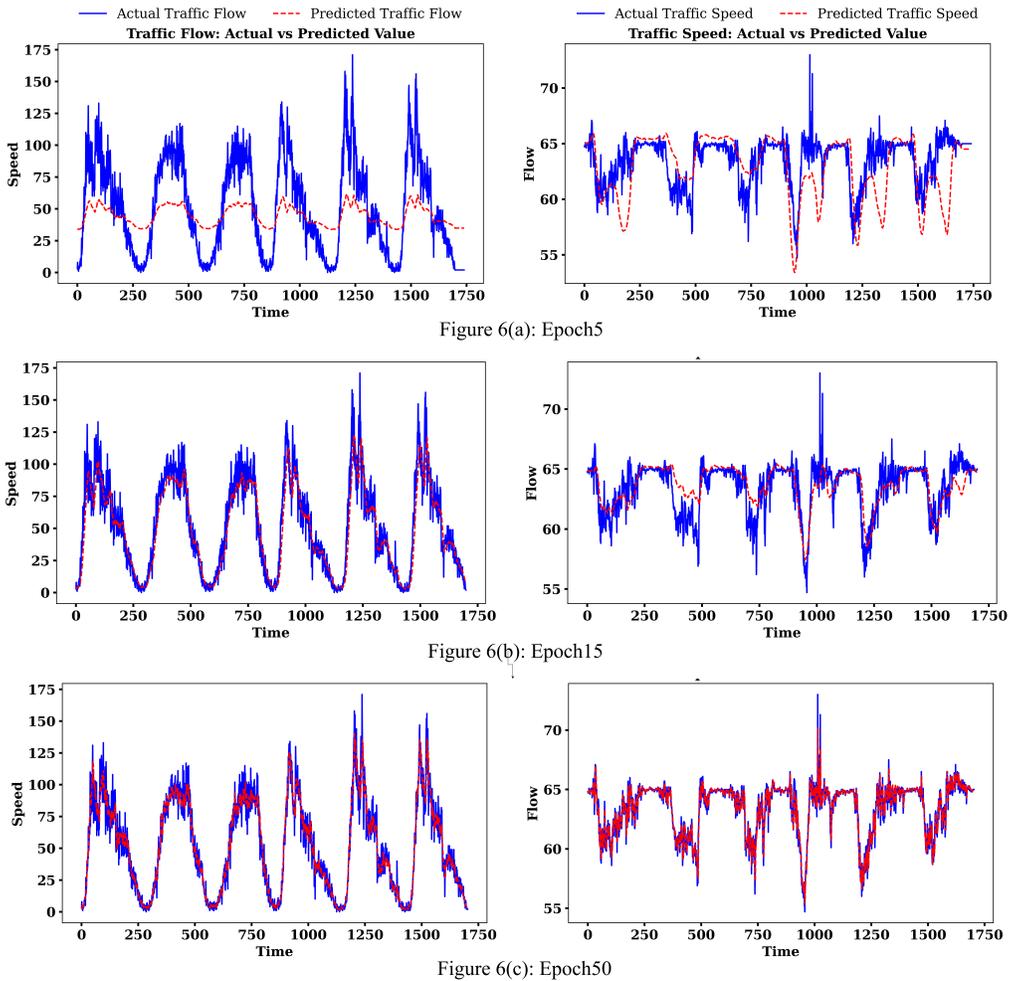}
\caption{Visualization of Traffic Flow and Speed Multi-Task Prediction at Epoch 5, 15, and 50}
\label{fig:image1}
\end{figure}

\subsection{Scalability considerations}

The scalability of our proposed DG-STMTL framework is a critical factor for its application to large-scale spatio-temporal tasks. To assess scalability, we have analyzed the computational complexity of the core modules (CTKE, HAMG, GSTGC) and their integration in Section 4.6. The overall time complexity of our model is primarily influenced by the operations within the GSTGC modules. This process scales with $O(N^2)$  due to the matrix operations. However, this computational cost is mitigated by leveraging sparse representations of the adjacency matrices, which significantly reduce the number of effective operations. From a memory perspective, the static matrix serves as a compact and computationally lightweight prior, which is fixed during training, ensuring minimal overhead regardless of dataset size. Similarly, the dynamically learned matrix is generated directly from task-specific data, making it adaptable to varying data characteristics while maintaining computational efficiency. The gating mechanism further supports scalability by selectively combining static and dynamic components, reducing redundancy in the hybrid adjacency matrix. Overall, the framework's design balances computational cost and adaptability, making it applicable to datasets of diverse scales and complexities. Future work may investigate further optimizations for extreme-scale datasets, including distributed training and hierarchical modeling.

\subsection{Practical Implications}

The proposed model offers substantial benefits for intelligent traffic management systems by enhancing the accuracy of spatio-temporal traffic predictions. This improvement enables city planners and traffic controllers to implement more effective traffic management strategies, such as adaptive traffic signaling and dynamic route optimization, leading to reduced traffic congestion and improved road safety. Beyond traffic management, our model has significant implications for any field that relies on spatio-temporal predictions. For example, in environmental monitoring, it can predict changes in air quality or weather conditions across different times and locations, aiding in more proactive responses to environmental hazards. In public safety, it can enhance surveillance systems by predicting crowd movements and potential security threats in real-time. Additionally, in urban planning, it assists planners in understanding population movements and urban dynamics, facilitating more informed decisions on infrastructure development and resource allocation.

\section{Conclusions}

This work introduces a novel graph-based framework for multi-task spatio-temporal prediction that effectively addresses the limitations of static adjacency matrices and inefficiencies in prior MTL strategies. By integrating a dynamic and task-specific gating system within our proposed CTKE unit, the model demonstrates superior ability to capture complex spatio-temporal relationships in multi-task settings. The designed GCN architecture and task-agnostic output layer further enhances the capability in handing STMTL tasks. Our experimental results confirm that the model outperforms existing baselines, achieving state-of-the-art performance on two real-world datasets. Given its adaptability and robust performance, we are confident that our model has the potential for broad application across various MTSTL domains, including but not limited to finance, climate science, and healthcare. 
Looking ahead, GCNs can still present challenges in interpretability, especially in complex settings. Future work will focus on enhancing the clarity of GCN models by leveraging advancements in explainable AI. Enhancing the transparency of GCN decision-making through explainable AI will be crucial, especially in fields where clear and trustworthy explanations are essential.


\bibliographystyle{elsarticle-num} 
\bibliography{ref}
\end{document}